# Perception-aware Tag Placement Planning for Robust Localization of UAVs in Indoor Construction Environments

Navid Kayhani [1], Angela Schoellig [2], Brenda McCabe [3]

[1] Ph.D. Candidate, Department of Civil and Mineral Engineering, University of Toronto, Toronto, ON, Canada M4S A4, Email: navid.kayhani@mail.utoronto.ca (corresponding author)

[2] Associate Professor, Institute for Aerospace Studies, University of Toronto, North York, ON, Canada M3H 5T6

[3] Professor, Department of Civil and Mineral Engineering, University of Toronto, Toronto, ON, Canada M4S 1A4

## ABSTRACT

Tag-based visual-inertial localization is a lightweight method for enabling autonomous data collection missions of low-cost unmanned aerial vehicles (UAVs) in indoor construction environments. However, finding the optimal tag configuration (i.e., number, size, and location) on dynamic construction sites remains challenging. This paper proposes a perception-aware genetic algorithm-based tag placement planner (*PGA-TaPP*) to determine the optimal tag configuration using 4D-BIM, considering the project progress, safety requirements, and UAV's localizability. The proposed method provides a 4D plan for tag placement by maximizing the localizability in user-specified regions of interest (ROIs) while limiting the installation costs. Localizability is quantified using the Fisher information matrix (FIM) and encapsulated in navigable grids. The experimental results show the effectiveness of our method in finding an optimal 4D tag placement plan for the robust localization of UAVs on under-construction indoor sites.

## INTRODUCTION

Autonomous mobile robots, including unmanned aerial vehicles (UAVs), are regarded as next-generation reality-capture technology for construction applications (Cai et al. 2019). When equipped with cameras, these platforms allow for the automated capture of high-quality images from user-specified locations and angles, which can significantly enhance the performance of downstream vision-based analytics (Hamledari et al. 2017). Robust localization is the first and foremost necessity to enable reliable autonomous navigation (Cadena et al. 2016). Localization is the problem of estimating a mobile system's pose, i.e., position and orientation, with respect to a reference frame. Robustness in localization refers to the method's ability to tolerate perturbations and challenging conditions while preserving the stability and integrity of the estimation. Thus, robust localization entails actively considering perception requirements and environmental information (Yi et al. 2019).

Robotic reality capture solutions deployed on outdoor sites rely mainly on the Global Navigation Satellite System (GNSS) signals for localization in autonomous mode (Martinez et al. 2021). In GNSS-denied indoor construction environments, most state-of-the-art autonomous platforms (Asadi et al. 2020; Ibrahim et al. 2019; Xu et al. 2019) rely on environment maps generated in advance by teleoperating the platform within the workspace. However, the layout changes with construction progress, requiring frequent costly and tedious mapping sessions. Additionally, common characteristics of indoor construction settings, such as untextured or repetitive areas (Xu et al. 2020), dynamic or temporary objects (Cadena et al. 2016), and reflective surfaces, reduce the quality of pre-built maps. This approach increases the likelihood of localization failure in indoor construction applications and reduces the system's robustness. Furthermore, the costs of the proposed custom-built prototypes (Asadi et al. 2020; Kim et al. 2018) and commercial products limit their scalability and applicability in practice.

To reduce the cost of automated indoor data collection and address the technical challenges discussed, we (Kayhani et al. 2022) previously proposed a low-cost, lightweight, versatile, tag-based visual-inertial localization method using AprilTags. Even though tags themselves have almost no cost, the manual tag installation process might become a tedious task if improperly planned (Kayhani et al. 2020). With changes in the interior layout, some tags may be occluded or need replacement, the target areas to be monitored or no-fly zones may vary, and tag placement options may change. Larger tags can be detected from longer ranges, whereas printing them on standard-sized sheets (e.g., letter or A4) may be more convenient and cost less. The number of tags must be kept at a minimum to limit the amount of manual work for placing and maintaining tags. Finally, the opted tag number, size, and location must result in robust and high-quality tag-based visual-inertial localization. These considerations highlight that experience-based tag placement planning is non-trivial, time-demanding, and may result in performance deviations.

The question arises about optimizing the tag placement/replacement process on an indoor construction site by maximizing localizability with a limited number of tags considering construction progress, project schedule, and safety requirements. Localizability, herein, refers to the agent's ability to reliably locate itself with low uncertainty and represents the estimation quality and robustness. This paper addresses the problem of four-dimensional (4D) tag placement, which aims to find the optimal tag configuration (i.e., number, size, and location) over time, considering the construction schedule and safety constraints (e.g., no-fly zones). The optimal solution maximizes the tag-based visual-inertial localization quality and certainty (i.e., tag-based localizability) while minimizing the installation costs. The pose estimation certainty can be quantified using the Fisher information (Zhang and Scaramuzza 2019). Therefore, the maximum tag-based localizability is achieved by devising a tag configuration with the highest expected information gained from tag measurements. Minimizing the installation costs is obtained by limiting the number of tag placements, replacements, and removals during the project.

Tag placement planning can be viewed as a constrained combinatorial optimization problem. It maximizes a localizability utility function within user-specified regions of interest (ROIs) while minimizing a cost function to penalize tag network modifications. ROIs include on-site target work zones and travel paths connecting these target locations with different importance levels. The utility function represents the likelihood of getting high-quality tag measurements by evaluating the attainable photometric information at each query point within ROIs. In other words, utility is determined not only by the tags' visibility but also by their informativeness. Minimizing the number of tag placements, replacements, and removal are obtained by associating a negative utility, i.e., cost, for adding, substituting, or removing tags in the network.

This paper addresses the gap in the literature by proposing a 4D-BIM-based solution for the automated design of tag placement plans, considering indoor construction dynamics. A perception-aware genetic algorithm-based tag placement planner (*PGA-TaPP*) is proposed to find the optimal tag configuration. It maximizes localizability while minimizing the total cost of tag installations by incorporating the project schedule and safety requirements. *PGA-TaPP* considers multiple project phases, tag placement heights, flight altitudes, and localizability metric functions. The proposed method inputs the project's 4D BIM, ROIs, no-fly zones, and user-specified design parameters. Each project comprises construction phases, representing project snapshots at different times. The site layout geometry of every phase is automatically extracted. Each phase is horizontally sliced to generate 2D scenes corresponding to different flight altitudes. Scenes, made up of polygons and ROIs, are further quantized to obtain discrete tag placement options and UAV query pose. The gainable information from query poses in the workspace is quantified using the Fisher information matrix (FIM). Finally, the developed genetic algorithm engine optimizes the tag configuration throughout the entire project such that (1) the overall tag-based localizability within ROIs is maximized; (2) the tag configuration adjustments are minimal; and (3) safety requirements are satisfied. The main contributions of this work are: (1) proposal of a perception-aware 4D tag placement planning method to find the optimal tag number, size, and location for supporting robust tag-based visual-inertial localization on indoor sites considering localizability, project schedule, installation costs, and safety; (2) quantification of attainable information for 6-DoF tag-based visual-inertial localization using Fisher information with various metric functions; and (3) quantitative evaluation of the performance of the proposed tag networks in a BIM-enabled simulation environment.

## BACKGROUND

### *Automated Data Collection Using Mobile Robots*

**Table 1 –** Automated data collection using mobile robots.

| Application Areas | Platform | Localization | Indoors | Application Examples |
|---|---|---|---|---|
| Infrastructure inspection | Aerial | GNSS | No | (Freimuth and König 2018) |
| Bridge inspection | Aerial | GNSS | No | (Lin et al. 2021) |
| Earthwork surveying | Aerial | GNSS | No | (Siebert and Teizer 2014) |
| Quality control | Aerial | GNSS | No | (Kielhauser et al. 2020) |
| Safety inspection | Aerial | GNSS | No | (Martinez et al. 2021) |
| Progress monitoring | Aerial + Ground | Pre-built maps | Yes | (Asadi et al. 2020) |
| Environmental air quality | Ground | Pre-built maps | Yes | (Jin et al. 2018) |
| Semantic modeling | Ground | Pre-built maps | Yes | (Adán et al. 2020) |
| Building retrofit performance | Ground | AprilTags | Yes | (Mantha et al. 2018) |

Frequent, reliable, and high-quality data from the job site are necessary for a systematic performance evaluation of architecture, engineering, and construction (AEC) projects (Moselhi et al. 2020). With advances in computer vision-based solutions, RGB images are considered one of the most valuable data modalities for automating inspection and monitoring tasks (Pal and Hsieh 2021). However, manual visual data acquisition is time-consuming, error-prone, and costly (Teizer 2015). In the past decade, automated data collection using autonomous mobile robots has gained momentum in the AEC community. Previous studies have extensively reviewed these applications (Cai et al. 2019; Ham et al. 2016; Rakha and Gorodetsky 2018). Outdoor visual data collection for infrastructure inspection (Freimuth and König 2018; Lin et al. 2021), earthwork surveying (Siebert and Teizer 2014), quality control (Kielhauser et al. 2020), safety inspection (Gheisari et al. 2014; Martinez et al. 2021), as well as indoor progress monitoring (Hamledari et al. 2017) are some examples of the extensive usage of UAVs in construction applications. However, UAVs in these studies were remotely controlled or relied on GNSS signals for autonomous flight. Autonomous ground robots were also deployed indoors for environmental air quality (Jin et al. 2018), construction progress monitoring (Asadi et al. 2020), semantic modeling (Adán et al. 2020), and building retrofit performance simulation (Mantha et al. 2018). Table 1 provides some examples of the applications of mobile robots in automated data collection in the AEC industry.

### *Indoor Localization in Construction Settings*

Indoor construction sites are unique GNSS-denied environments. These settings present technical difficulties for many localization methods. Wave-based methods such as wireless local area networks (WLAN) (Deasy and Scanlon 2004), radio-frequency identification (RFID) (Liu et al. 2014; Razavi and Moselhi 2012), and ultra-wideband (UWB) (Witrisal and Meissner 2012), lose their accuracy and are unreliable in indoor construction settings due to interference with construction materials (e.g., steel) (Ibrahim and Moselhi 2016). A common technique for navigating unknown workspaces is to concurrently map the environment and localize the agent within the built map, referred to as simultaneous localization and mapping (SLAM). Modern SLAM architectures constitute a back-end state estimation component supported by a sensor-dependent front-end for feature extraction, data association, and loop closure (i.e., recognizing a previously observed place) (Cadena et al. 2016). Loop closure is necessary to reset the localization error and estimate the actual topology of the environment in the map. SLAM reduces to odometry without loop closure, which drifts over time and is unreliable for long-term navigation.

In recent decades, significant research has been conducted to improve localization and state estimation accuracy and efficiency in GNSS-denied environments, resulting in promising outcomes (Delmerico and Scaramuzza 2018). However, these results have been demonstrated only for particular domains and given a hardware-environment-performance combination (Cadena et al. 2016). For instance, successful SLAM/mapping missions with sufficient accuracy (<10 cm) in an office building involving a ground robot with wheel encoders, a laser scanner, and enough computational and power resources are achievable (Mur-Artal et al. 2015). However, performing SLAM/mapping may fail in highly dynamic environments with perceptual aliasing and feature scarcity (Cadena et al. 2016) and is considered resource-intensive for large environments (Muñoz-Salinas et al. 2019).

### *Tag-based Visual-Inertial Localization*

Fiducial markers such as AprilTags (Wang and Olson 2016) are planar artificial landmarks consisting of patterns. They provide robust data association and are ideal for featureless or repetitive areas common in construction settings. Tags are often employed for local positioning tasks such as map initialization and UAV landing (Brommer et al. 2018). For long-term localization, visual-SLAM based on only tags (Muñoz-Salinas et al.

2019) and coupled with keypoints (Muñoz-Salinas and Medina-Carnicer 2020) were proposed. Relying only on vision reduces their reliability in the face of occlusion or motion blur. Tag-based visual-inertial localization (Kayhani et al. 2022) is an inexpensive alternative to enable autonomy for low-cost UAVs to be deployed for automated data collection in indoor construction settings. This method uses tags with known sizes and locations, jointly acting as a quasi-map of the environment. The 3D position accuracy of as low as $2 - 5\ cm$ was obtained in the experiments conducted in the laboratory and simulation settings. Our proposed formulation is based on an on-manifold EKF, suitably addressing the topological structure of the rotation and pose groups in 3D. It integrates two sources of information to yield consistent 6-DOF global pose estimates in real-time: (1) odometry-based rotational and translational velocities and (2) tag-based visual measurements.

$$\text{Motion model} \qquad \mathbf{x}_k = f(\mathbf{x}_{k-1}, \mathbf{u}_{k-1}) + \mathbf{w}_{k-1} \qquad (1)$$

$$\text{Measurement model} \qquad \mathbf{y}_k = g(\mathbf{x}_k) + \mathbf{n}_k \qquad (2)$$

An EKF involves two steps as a recursive filter: prediction and correction. The prediction step propagates the current state mean and covariance estimates forward in time. The current pose $\mathbf{x}_k$ is predicted based on the previous pose estimate $\mathbf{x}_{k-1}$ and the odometry-based velocities (inputs) $\mathbf{u}_{k-1}$ considering a zero-mean Gaussian noise $\mathbf{w}_{k-1}$ through a motion model ($f(.)$) (See Eq. 1). The predictions are updated in a correction step where tag measurements $\mathbf{y}_k$ are incorporated, and the posterior mean and covariance are estimated. The correction is based on a measurement model ($g(.)$), perturbed by a zero-mean Gaussian noise ($\mathbf{n}_k$) (Eq. 2). In a related study (Neunert et al. 2015), a tag-based visual-inertial EKF-SLAM was proposed. However, their proposed filter-based SLAM can only map a limited number of tags in real-time, limiting its application on large construction sites.

### *Tag-based Localization Quality and Fisher Information*

Considering the reliability of autonomous navigation at the planning stage prioritizes tag configurations that can result in high-quality localization. The observedtag's information must be quantified to incorporate the knowledge about tag-based visual-inertial localization quality. Fisher information is often used to represent this knowledge (Zhang and Scaramuzza 2019). According to the Cramér-Rao lower bound (CRLB), the covariance of any unbiased estimate $\mathbf{x}$, given a set of measurements $\mathbf{y}$, is bounded by the inverse of the Fisher information matrix (FIM) (Barfoot 2017), denoted as $\mathbf{I}_{\mathbf{x}}(\mathbf{y})$. In other words, the measurement information about the parameter to be estimated establishes a fundamental limit on how confident the estimates can be, regardless of the form of the estimator, and the FIM quantifies this notion. Because localization is essentially a pose estimation problem, the FIM can be applied to quantify the estimation uncertainty and, therefore, viewed as a metric for localization quality quantification.

Given that the measurement process of the state, i.e., 6-DoF UAV's pose, can be described as a likelihood function $p(\mathbf{y}|\mathbf{x})$, the observed Fisher information can be defined as (Barfoot 2017):

$$\mathbf{I}_\mathbf{x}(\mathbf{y}) = E\left[\left(\frac{\partial \ln p(\mathbf{y}|\mathbf{x})}{\partial \mathbf{x}}\right)^T \left(\frac{\partial \ln p(\mathbf{y}|\mathbf{x})}{\partial \mathbf{x}}\right)\right] \tag{3}$$

Assuming zero-mean Gaussian noise with constant covariance $\mathcal{N}(\mathbf{0}, \mathbf{\Sigma})$ for measurements, Eq. 3 can be rewritten as (Wang and Dissanayake 2008):

$$\mathbf{I}_\mathbf{x}(\mathbf{y}) = (\mathbf{G}_\mathbf{x})^T \mathbf{\Sigma}^{-1} \mathbf{G}_\mathbf{x} \tag{4}$$

where $\mathbf{G}_\mathbf{x}$ is the Jacobian of the measurement model in Eq. 2:

$$\mathbf{G}_\mathbf{x} = \frac{\partial g}{\partial \mathbf{x}} \tag{5}$$

## METHODS

### Overview

In this work, 4D-BIM is utilized to retrieve the project's spatial-temporal information and incorporate the dynamic nature of indoor construction environments. As illustrated in Figure 1, the proposed method inputs the project 4D-BIM in Autodesk *Revit*® or Industry Foundation Classes (IFC) format to address the ease of design and facilitate interoperability. The 4D-BIM includes the 3D model, project schedule, and ROIs for the time-stamped construction phases. Other inputs include the UAV's camera specifications, tag information, and design factors such as planning and optimization parameters.

The *Initialization and Geometry Extraction* module automatically extracts the type, geometry, positioning, and relationship of the relevant elements for each time-stamped construction phase, converts them to a set of 2D polygons, and stores them as separate files. These files are then fed into the proposed perception-aware genetic algorithm-based tag placement planner (*PGA-TaPP*) to find the optimum tag configuration that supports robust tag-based visual-inertial localization, considering the project progress and schedule. The extracted geometry of the site layout, ROIs for each phase, and input parameters are fed to the *Modeling and Problem Formulation* module. These parameters include the UAV's camera specifications (e.g., depth of view (DoV), lens parameters, and vehicle-to-camera transform), tag information (e.g., available sizes, installation heights, maximum number), and planning parameters (e.g., flight altitudes, discretization resolutions, and no-fly zones). A set of 2D scenes corresponding to each flight altitude is first generated, each of which comprises polygons representing modified ROIs and building elements. The polygons are further quantized to obtain grids, i.e., discretized navigable areas and tag placement locations. The discretized navigable areas constitute constraint-free grid cells within ROIs at each flight altitude. Tag placement options are a discrete set of feasible tag installation locations in each phase.

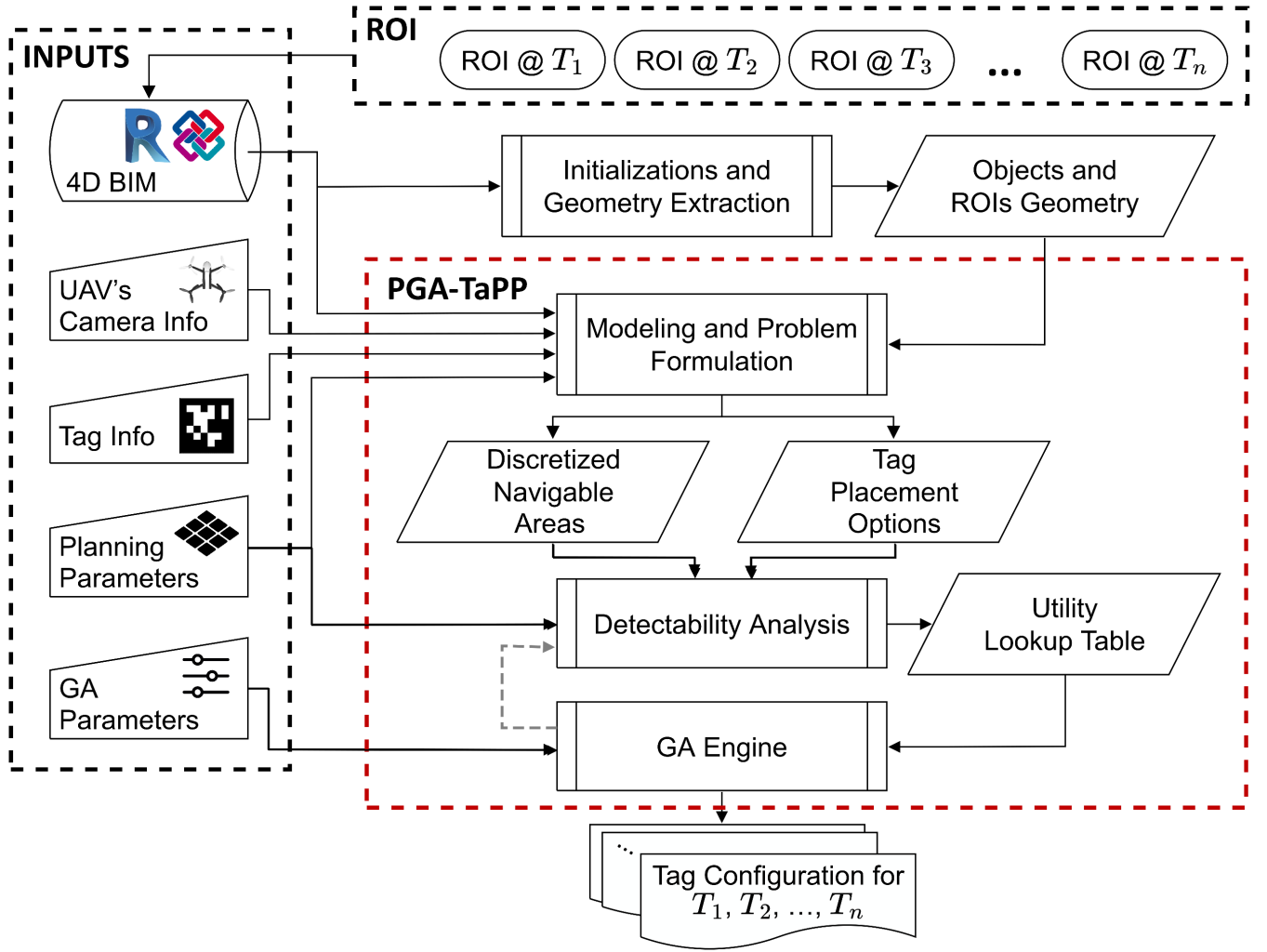

**Figure 1-** Overview of the proposed method.

To calculate the utility function for a solution (i.e., tag placement network) in a particular time-stamped construction phase, *PGA-TaPP* quantifies the quality of expected tag measurements in discretized UAV orientations within every ROI cell. It first identifies the tags visible from each pose and assigns a utility to the given query pose. The utility of the query pose is calculated separately for each visible tag and is a function of the FIM (more discussion in *Scoring Metric Functions for Localizability Evaluation*). A lookup table is then constructed to efficiently search the solution space based on the identified discretized navigable areas, the tag placement options, and the planning parameters. The created lookup table allows for efficient querying of the utility for any UAV pose and tag placement option. The table can be constructed entirely prior to GA-based optimization or as the optimization process is performed. In the former mode, the *Detectability Analysis* module extensively explores the tag placement possibilities for efficient exploitation in searching the solution space by the optimization module, denoted as *GA Engine*. Finally, the *GA Engine* finds the tag configuration for each construction phase by maximizing the vehicle's localizability with minimal tag configuration modifications throughout the project. The resulting tag configuration for each time-stamped construction phase is visualized for enhanced communication.

***Initializations and Geometry Extraction***

The *Initializations and Geometry Extraction* module takes the project's 4D-BIM model as input. This module uses an Application Programming Interface (API) to handle *Revit®* files and relies on open-source libraries to read IFC files in the back-end. Supporting both formats is essential because IFC is a neutral schema that allows interoperability (Laakso 2012), yet arguably more suitable for information retrieval than design. In contrast, users may need to modify the planning constraints (e.g., ROIs or no-fly zones) or incorporate potential discrepancies between BIM and the actual site conditions throughout the project. The discrepancies can be caused by modeling errors, infrequent 4D-BIM updating, or temporary site objects. These changes can be made more conveniently in a modeling tool with a graphical user interface (e.g., Autodesk *Revit®*).

The project is first divided into time-stamped construction phases (i.e., $T_1, T_2, \dots, T_m$) according to a given data collection mission plan. The algorithm checks if the BIM file has a *Project Base Point* to set the origin of the global coordinate system. If no base point is found, an arbitrary but consistent base point is chosen, e.g., the bottom left corner of the building envelope. For each construction phase, i.e., $T_i$ ($i \in [1,2, \dots, m]$), it automatically selects the relevant building elements (e.g., walls, partitions, doors, windows, columns), ROIs, and no-fly zones. The indoor site layout at $T_i$ is further divided into discretized altitudinal layers to account for the UAVs' 3D maneuver and its impact on camera FOV and tag detectability (Figure 2). The navigable altitude of the UAV might also change with construction progress. For example, installing suspended ceilings or electrical fixtures reduces the maximum safe flight altitude (Hamledari et al. 2021). Thus, the unnavigable areas at each altitudinal layer are added to no-fly zones at the corresponding altitude. Finally, the indoor site layout at each layer is reduced to a set of 2D polygons, including building elements, ROIs, and no-fly zones.

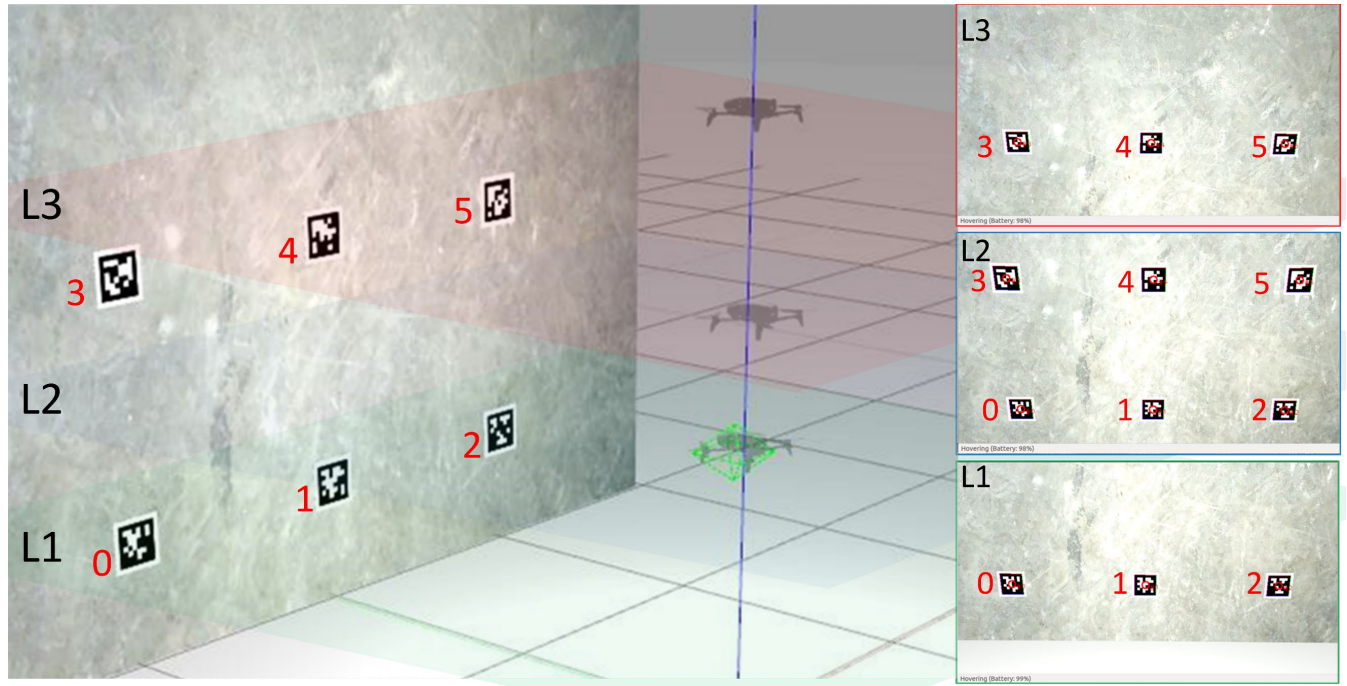

**Figure 2 -** The altitudinal layers and the detected tags in the image stream captured at each flight altitude.

### *PGA-TaPP: Perception-aware Genetic Algorithm-based Tag Placement Planner*

*PGA-TaPP* involves four main steps: (1) camera and scene modeling; (2) optimization formulation and localizability metric establishment; (3) detectability analysis and utility calculations: and (4) GA-based optimization. The first two steps are included in the *Modeling and Problem Formulation* module, and *Detectability Analysis* and *GA Engine* handle the rest.

**Modeling and Problem Formulation**

UAV specifications, tag information, planning parameters, and the extracted geometric information from building elements, ROIs, and no-fly zones are fed to the first module in *PGA-TaPP* for scene modeling and problem formulation. This section discusses the module's components: camera and measurement models, scene modeling, and scoring metric formulation.

*Camera and Measurement Models*

The UAV's camera is represented as a pinhole camera model with given intrinsics (e.g., optical center, focal length, and field of view) and the camera to vehicle transformation. This allows for covering different robotic platforms with arbitrary camera configurations. Without loss of generality, it is assumed that the UAV has only one camera on board for the rest of this paper. The 3D coordinates of the $n$-th corner point (e.g., $n \in \{1, 2, 3, 4\}$ for square markers) of tag $j$ expressed in the camera frame, $\mathbf{p}_c^{p_{\tau_j,n^c}}$, can be written as (Kayhani et al. 2022) :

$$z^{\tau_j,n}(\mathbf{x}) = \mathbf{p}_c^{p_{\tau_j,n^c}} = \mathbf{D}^T \mathbf{T}_{cv} \mathbf{T}_{vw} \mathbf{p}_{\tau_j,n}^w = [X \quad Y \quad Z]^T \tag{6}$$

where $\mathbf{p}_{\tau_j,n}^w$ is the homogenous coordinate of the $n-th$ corner point of tag $j$ in the world frame, and $\mathbf{T}_{vw} = \{\mathbf{r}_w^{vw}, \mathbf{C}_{vw}\} \in SE(3)$ is the vehicle's pose expressed in the world frame and the state to be estimated ($\mathbf{x}$). Additionally, $\mathbf{T}_{cv}$ is the vehicle-to-camera transformation known from calibration, and $\mathbf{D}^T = [\mathbf{1}_3 | \mathbf{0}_{3 \times 1}]$ is a dilated identity matrix to refine the matrix dimensions.

Denoting the pinhole camera model that projects $\mathbf{p}_c^{p_{\tau_j,n^c}}$ into a rectified image as $s(.)$, the measurements, i.e., the pixel coordinates of the $n$-th corner point of tag $j$, are obtained as (Kayhani et al. 2022):

$$g^{\tau_j,n}(\mathbf{x}) = s\big(z^{\tau_j,n}(\mathbf{x})\big) = \begin{bmatrix} u \\ v \end{bmatrix} = \mathbf{D_p} \begin{bmatrix} f_u & 0 & c_u \\ 0 & f_v & c_v \\ 0 & 0 & 1 \end{bmatrix} \frac{1}{Z} \begin{bmatrix} X \\ Y \\ Z \end{bmatrix} \tag{7}$$

where $g^{\tau_j,n}(\mathbf{x})$ is the noise-free measurement model, $c_u$ and $c_v$ are the optical offsets (principal point), $f_u$ and $f_v$ are the camera focal lengths, and $\mathbf{D_p} = [\mathbf{1}_2 | \mathbf{0}_{2 \times 1}]$ is a dilated identity matrix.

Before localizability quantification for a pose, it is essential to identify the tags detectable in the image captured by the vehicle at that query pose. A tag can be detected from $\mathbf{T}_{vw}$ if (1) it can be observed from the vehicle position $\mathbf{r}_w^{vw}$, considering occlusions; and (2) it is within the camera field of view (FoV). The former is obtained using visibility analysis and generating a visibility graph (Lozano-Pérez and Wesley 1979) to identify visible tag corners ($\mathbf{P}^{\tau_j w}$). The latter is achieved by projecting the tag corners onto image plane (Eq. 7) and checking whether the projected pixel coordinates $\boldsymbol{\pi} = [u, \ v]^T$ are within the image boundary $I$ (Eq. 8). More details are provided in the *Detectability Analysis* section.

$$v(\mathbf{T}_{vw}, \mathbf{P}^{\tau_j w}) = \begin{cases} 1, & \boldsymbol{\pi} \in I \\ 0, & \boldsymbol{\pi} \notin I \end{cases} \tag{8}$$

*Scene Modeling*

As shown in Figure 3, a project is divided into phases, corresponding to snapshots of the project. Accounting for the 3D maneuver of UAVs, each phase is subdivided into altitudinal layers (Figure 2), considering the safety of operation. Accordingly, a 2D scene is generated for each altitudinal layer to model the given indoor site at time $T_i$. Each scene initially consists of 2D polygons representing building elements, ROIs, and no-fly zones. The building elements collectively correspond to inside and outside boundaries, obstacles, and surfaces for tag installation. The ROI polygons specify the geometry within which the localizability must be maximized, whereas no-fly zones represent where the UAV is prohibited from operating. The difference between ROI and no-fly zone polygons is found for each scene to find the navigable area, denoted as modified ROIs.

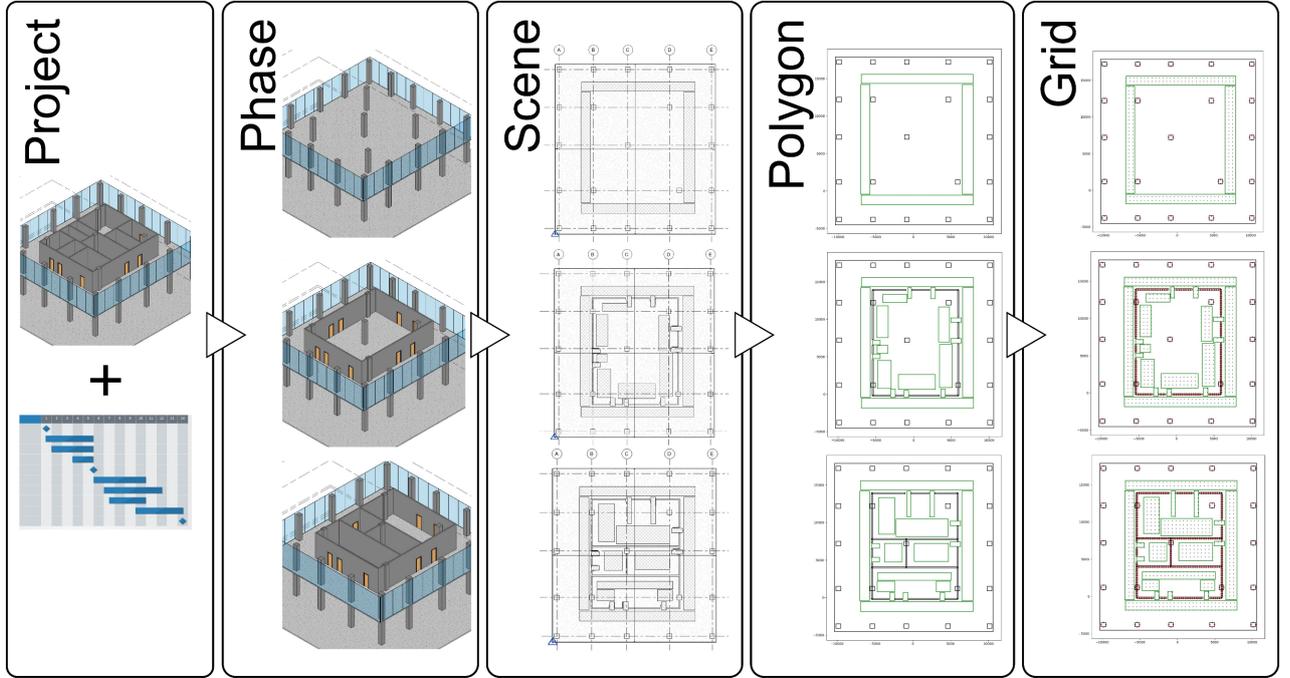

**Figure 3 -** Hierarchy of the spatial objects in *PGA-TaPP*.

---

**Algorithm 1** Tag Option Identification

---

1: INPUT $obs$, $d_{res}$, $t_{size}$            ▷ $obs$: obstacles
2: RETURN List$[x, y, u, v]$            ▷ tag direction = $[u, v]$
3: $dist = $ FindMinimumTagDistance($d_{res}, t_{size}$)
4: $\epsilon = 5 \times 10^{-5}$
5: **for** $poly$ in $obs$ **do**
6:      $poly = $ Scale($poly$, $1+\epsilon$)      ▷ to avoid overlaps with surface
7:      $lines = $ DecomposePoly($poly$)      ▷ list of surfaces
8:      **for** $line$ in $lines$ **do**
9:          $x, y = $ InterpolateByDistance($line$, $dist$)
10:          $u, v = $ GetNormal($line$)
11:          List.append($x, y, u, v$)
12:      **end for**
13: **end for**
14: return List

---

**Figure 4 -** Algorithm for identifying quantized tag placement options in 2D.

Modified ROI polygons are then quantized into grid cells. The default grid cell size is set to $0.5\ m \times 0.5\ m$, considering compact UAVs' dimensions plus a clearance buffer to guarantee safety. The center of the grid cells constitutes the query points for evaluating localizability. We note that the query point corresponds to the vehicle position, and the camera pose is obtained using the inputted vehicle-to-camera transform ($\mathbf{T}_{cv}$). Our approach finds the optimal tag network from a discrete space of possible tag placement locations. Tag placement options are determined based on user-specified installable surfaces, discretization resolution, and tag size(s) such that tags do not overlap. Considering practicality, economy, and ease of installation, the potential tag placement heights are limited to those at which tags can be installed without extra apparatus and additional labor input. The algorithm for finding quantized tag placement options in 2D is summarized in Figure 4. The 2D options found are then added to each level to consider multi-level tag placement (Figure 2).

*Fisher Information Matrix for Tag-Based Visual-Inertial Localization*

This section focuses on formulating the Fisher information for a single pose in on-manifold tag-based visual-inertial localization. Calculating the FIM (Eq. 4) for a single pose $\mathbf{x} = \mathbf{T}_{vw}$ requires calculating the Jacobian of the observation of $n - th$ corner point of tag $j$ as in Eq. 9. The Jacobian in Eq. 9 is calculated using two factors defined in Eq. 10 and Eq. 11. These factors come from the introduced nonlinearities in the measurement model (Eq. 7) linearized about their mean (for more details on Jacobian derivations see (Kayhani et al. 2022)).

$$\mathbf{G}^{\tau_{j,n}} = \frac{\partial g^{\tau_{j,n}}(\mathbf{x})}{\partial \mathbf{x}} = \frac{\partial}{\partial \mathbf{x}}\left(s\left(z^{\tau_{j,n}}(\mathbf{x})\right)\right) = S^{\tau_{j,n}} Z^{\tau_{j,n}} \tag{9}$$

$$S^{\tau_{j,n}} = \mathbf{D}_{\mathrm{p}} \begin{bmatrix} f_u & 0 & c_u \\ 0 & f_v & c_v \\ 0 & 0 & 1 \end{bmatrix} \begin{bmatrix} \dfrac{1}{Z} & 0 & -\dfrac{X}{Z^2} \\ 0 & \dfrac{1}{Z} & -\dfrac{Y}{Z^2} \\ 0 & 0 & 0 \end{bmatrix} \tag{10}$$

$$Z^{\tau_{j,n}} = \mathbf{D}^T \mathbf{T}_{cv} \left(\mathbf{T}_{vw}\mathbf{p}^w_{\tau_{j,n}}\right)^{\odot} \tag{11}$$

Calculating $Z^{\tau_{j,n}}$ in Eq. 9 requires introducing $(.)^{\odot}$ operator (Barfoot 2017) that acts on $4 \times 1$ points in homogenous coordinates following Eq. 12, where the first three components ($\boldsymbol{\varepsilon} \in \mathbb{R}^3$) are expressed in the skew-symmetric matrix format (Eq. 13).

$$\mathbf{p}^{\odot} = \begin{bmatrix} sx \\ sy \\ sz \\ s \end{bmatrix}^{\odot} = \begin{bmatrix} \boldsymbol{\varepsilon} \\ \eta \end{bmatrix}^{\odot} = \begin{bmatrix} \eta\mathbf{1} & -\boldsymbol{\varepsilon}^{\wedge} \\ \mathbf{0}^T & \mathbf{0}^T \end{bmatrix} \in \mathbb{R}^{4 \times 6} \tag{12}$$

$$\boldsymbol{\varepsilon}^{\wedge} = \begin{bmatrix} \varepsilon_x \\ \varepsilon_y \\ \varepsilon_z \end{bmatrix}^{\wedge} = \begin{bmatrix} 0 & -\varepsilon_z & \varepsilon_y \\ \varepsilon_z & 0 & -\varepsilon_x \\ -\varepsilon_y & \varepsilon_x & 0 \end{bmatrix} \tag{13}$$

The FIM for tag $j$ observed at $\mathbf{T}_{vw}$ is calculated using Eq. 4 and denoted as $\mathbf{I}_{\tau_j}(\mathbf{T}_{vw})$. Let $\mathcal{T} \subset \{\tau_1, \tau_2, \dots, \tau_M\}$ be the set of all tags detected in the image taken at the same vehicle pose, we have:

$$\text{FIM}(\mathbf{T}_{vw}) = \sum_{\tau_j}^{\tau_j \in \mathcal{T}} \mathbf{I}_{\tau_j}(\mathbf{T}_{vw}) \tag{14}$$

*Scoring Metric Functions for Localizability Evaluation*

The FIM is incorporated in the proposed optimization formulation to consider localization quality and quantify vehicle localizability. In any GA-based optimization, a mechanism is required to determine the utility (i.e., fitness) of the generated solutions (i.e., population), such that those with higher values survive and others are eliminated (i.e., selection). Therefore, the information matrix is mapped to a single scalar using a non-negative metric function $\mathcal{M}(.)$. This work incorporates the log-determinant, the trace, and the smallest eigenvalue as metric functions directly obtained from the FIM. Some fundamental properties of these metric functions are briefly reviewed to provide more intuition. The FIM determinant can be a proper information measure because its absolute value geometrically represents the volume of square matrices in $n$ dimensions. The larger the volume, the more informative the measurement will be in the case of the FIM. In this work, the transformed logarithm of the determinant is used to yield more numerical stability. The trace of a matrix is equal to the sum of its eigenvalues, and eigenvalues describe how a linear transformation stretches the space in a particular direction. Maximizing trace, also known as A-optimality, guarantees that the majority of the state space dimensions are considered. In a similar sense, the smallest eigenvalue represents the least affected direction. It corresponds to E-optimality, aiming to improve the worst-case variances of the parameter set by maximizing the smallest eigenvalue of the FIM.

**Detectability Analysis**

To quantify the expected attainable information from a query pose $\mathbf{P}_q(x_q, y_q, z_q, \theta_q)$, it is essential to predict the detectable tags from that pose. The necessary condition for a tag to be detectable is its visibility from the query pose. For square tags such as AprilTags, it is required to have all four corners in view to guarantee visibility (Kayhani et al. 2020), i.e., corners are within the camera's FoV and not occluded. Occlusion is found based on the camera line of sight, depending only on the camera and tag position. However, being within FoV depends on the vehicle's pose and camera intrinsics. A minimum side length threshold ($SL_{min}$) is defined to reliably predict the detectability of tags. The AprilTag detection algorithm (Wang and Olson 2016) has a pre-set threshold for rejecting small line segments (i.e., 4 pixels) in its line fitting module to minimize the chance of false-positive tag detections. It was experimentally observed that the minimum side length of the tag for reliable detections in practice is 15-20 pixels (Kayhani et al. 2019).

---
**Algorithm 2** Tag Detectability and FIM Calculation
---

1: INPUT $p_q(x, y, z)$, $\mathbf{p}_q(p_q, \theta)$, $scene$, $t_{locID}$     ▷ $t_{locID}$: tag's location ID
2: RETURN **FIM**
3: $UAV = scene.\text{uav}$, $Camera = UAV.\text{camera}$, $T_{cv} = UAV.\text{Tcv}$,
4: $DoV = Camera.\text{dov}$ $K = Camera.\text{k}$     ▷ $K$: camera intrinsics
5: **FIM** = ZEROS(6,6)
6: $p_c = \text{FINDTAGCENTER}(t_{locID})$
7: $p_1, p_2, p_3, p_4 = \text{FINDTAGCORNERS}(t_{locID}, t_{size})$
8: **if** DIS($p_c, p_q$) < $DoV$ **then**
9:     **for** $i$ in $[1, 2, 3, 4]$ **do**
10:        **if** !(LEEVISIBILITY($p_i, p_q$), INFOV($p_i, \mathbf{p}_q, K$)) **then**
11:           return **FIM**
12:        **end if**
13:     **end for**
14:     $L_{min} = \text{MINPROJECTEDLENGTH}(camera, p_1, p_2, p_3, p_4, p_q)$
15:     **if** $L_{min} > SL_{min}$ **then**
16:        **for** $i$ in $[1, 2, 3, 4]$ **do**
17:           **FIM** += CALCULATEFIM($T_{cv}, K, p_i, p_q$)     ▷ Eq. 9-13
18:        **end for**
19:     **end if**
20: **end if**
21: return **FIM**

---

**Figure 5 -** Algorithm for tag detectability analysis and FIM calculations for a single UAV pose and tag.

## GA Engine

Finding the 4D tag configuration that maximizes the localizability with the lowest installation costs, considering indoor site dynamics, is a constrained combinatorial optimization and non-deterministic polynomial-time (NP)–hard problem. A genetic algorithm (GA) is adopted in this work to tackle this optimization problem due to its advantages: (1) it can identify near-optimum solutions; (2) it has a low tendency to get stuck at local optima due to mutation; (3) it could be parallelized for efficient concurrent computations; and (4) it can handle multi-objective fitness functions and flexible policies. The GA-based optimization in *PGA-TaPP* is handled by an engine (Figure 7), developed with a modular architecture and parallel computation capabilities. The latter is a highly desirable feature in GA-based optimization as the fitness function calculations are often the bottleneck. The former allows for deploying different fitness functions and GA operators such as selection, mutation, and crossover.

The first requirement for employing GA-based optimization is defining a genetic representation of a feasible solution to the problem, i.e., chromosomes. In this work, chromosomes are structured as a sequence of elements (i.e., genes) representing the collection of all tag placement options in all construction phases (Figure 6). Each gene is associated with a global identifier, i.e., tag location ID ($ij$), unique among construction phases. A location ID $ij$ corresponds to tag location $i$ in construction phase $T_j$, which can point to a feasible or infeasible tag placement option at $T_j$ depending on the site layout. For example, if a wall is built at $T_m$, then the tag options corresponding to that wall are infeasible for the previous phases. Genes can take any non-negative integer value between zero and the number of tag sizes ($\tau_{ij} \in [0, n_s]$). Zero corresponds to no tags placed at the corresponding location, while any integer greater than zero embodies the installation of a particular tag size. If only a single tag size is available on site, the chromosomes will reduce to binary.

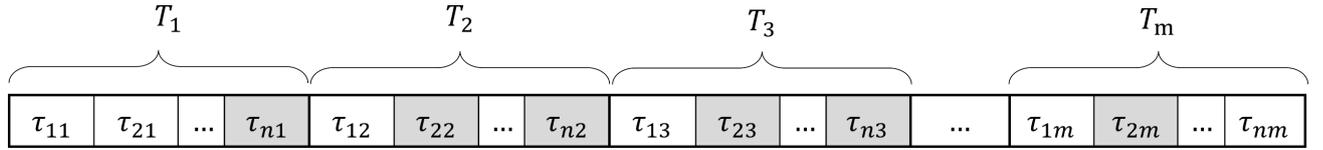

**Figure 6 –** Schematic chromosome structure containing tag placement options for a project with $\boldsymbol{m}$ phases (gray genes indicate the infeasible tag placement option at each construction phase).

To determine the best solution (i.e., fittest chromosome) in each iteration (i.e., population), a mapping (i.e., fitness function) is needed to assign a score (i.e., fitness score) to the generated solutions by the GA. The fitness score for a given chromosome is calculated based on a utility ($U$) that maximizes localizability and a cost that minimizes the installation costs by penalizing changes in the network in and between phases (Eq. 15).

$$Score = U - J \qquad (15)$$

For any given tag configuration, the grid-wise utility matrix for $roi \in ROI$ in phase $T_i$ is denoted as $\boldsymbol{U_{roi}}$ and obtained using Eq. 16. This requires estimating the expected FIM for each query cell $c_q(x_q, y_q)$ and mapping the obtained matrix value to a scalar using scoring metric function $\mathcal{M}(.)$. Estimating the FIM for $c_q(x_q, y_q)$ involves determining the FIM (Figure 5) for all discrete query poses in each scene $s \in S$, corresponding to the altitudinal layer $z_q$, with discrete yaw angles $\theta_q$ quantized by a user-specified rotational step ($\Delta\theta$) (e.g., 5°). By calculating the grid-wise utility, the algorithm respects the spatial structure of the grid cells. Finally, the phase $U_{T_i}$ and the total utility $U$ is found by aggregation using Eq. 17, where $I_{roi}$ is a relative importance factor associated with $roi$.

$$\boldsymbol{U_{roi}} = \sum_{z_q}^{s \in S} \sum_{\theta_q}^{\theta_q \in \theta} \mathcal{M}\big(\text{FIM}(x_q, y_q, z_q, \theta_q)\big), \qquad \forall c_q(x_q, y_q) \in roi \qquad (16)$$

$$U = \sum_{i=1}^{m} U_{T_i}, \qquad U_{T_i} = \sum_{roi}^{roi \in ROI} I_{roi} \times sum(\boldsymbol{U_{roi}}) \tag{17}$$

Utility calculations can be coupled or decoupled with the main GA optimization flow. When coupled, the tag placement options are explored as the current chromosome is evaluated (Figure 7). The FIM for each query pose and tag location ID pair is calculated once in exploration and stored in a hash lookup table for efficient exploitations. The entire tag options in $T_i$ can be exhaustively explored prior to optimization as an alternative. In both cases, parallel processing techniques were used to increase exploration efficiency. However, the decoupled approach allows for normalizing the gird-wise utility by the maximum utility obtainable from the grid, enhancing score interpretability and facilitating cost parameter assignments. The maximum utility is obtained when all tag placement options are occupied. Normalizing the grid-wise utility brings the utility between zero and one, represents the utilized capacity of each cell, and allows for easier comparisons and interpretability.

$$J = w_{plc} \left[ \sum_{i=1}^{n_s} \frac{n_{plc_i}}{\alpha_i} + \frac{n_{rmv}}{\lambda_{rmv}} + \lambda_{rpl} n_{rpl} \right], \qquad 0 < \alpha_i, \lambda_{rmv}, \lambda_{rpl} \leq 1 \tag{18}$$

$$w_{plc} = S_{min} \times P_c \times n_{cells} \tag{19}$$

The cost function $J$ penalizes adding extra tags to the network by expecting a minimum utility contribution. It also accounts for the sequential relation between construction phases by considering the entire project (i.e., chromosome) and penalizing the unnecessary changes in the tag network over time. The cost is a function of the number of placements ($n_{plc}$), replacements ($n_{rpl}$), and removals ($n_{rmv}$) throughout the entire project (Eq. 18). It assumes that tags need replacements every $k$ phases (e.g., due to damages or being covered), and the number of available tag sizes is $n_s$. An accessibility coefficient ($0 < \alpha_i \leq 1$) is defined for each tag size to account for its practicality. The tag size with the unit accessibility coefficient is denoted as the project reference tag. The minimum utility contribution of a single reference tag is denoted as $w_{plc}$ while $\lambda_{rmv}$ and $\lambda_{rpl}$ represent the relative weights for tag removals and replacements, respectively. Thanks to the normalized grid-wise utility calculations, $w_{plc}$ can be specified as the minimum score ($S_{min}$) that $P_c$ percent of the cells are expected to gain (Eq. 19).

As depicted in Figure 7, GA-based optimization begins with reading input data, including tag placement options, navigable grid cells, and visibility graphs for senses in each construction phase. Then, a random population of chromosomes is generated for all tag placement options across all construction phases. Next, the activated tag location IDs for each chromosome, i.e., non-zero genes, are identified. The utility is directly looked up from the hash table if they are already explored. Otherwise, it explores the scenes of all construction phases to obtain the utilities while updating the table for later exploitations. The fitness score for each chromosome in the population is calculated by incorporating the cost. The fittest pair is found and added to the next generation by sorting the population by score. The next generation's population is created by performing selection, crossover, and mutation operations. The selection operator randomly selects a pair of chromosomes whose fitness values weigh the selection probability. The crossover switches segments of the genes in selected pairs to create a new offspring based on different techniques (e.g., one-point, two-point, and uniform). The mutation creates a minor random tweak in the offspring chromosomes (e.g., shuffling and flipping) to introduce diversity in the population. Finally, the fittest chromosomes are returned if the termination condition is met (e.g., the maximum number of iterations).

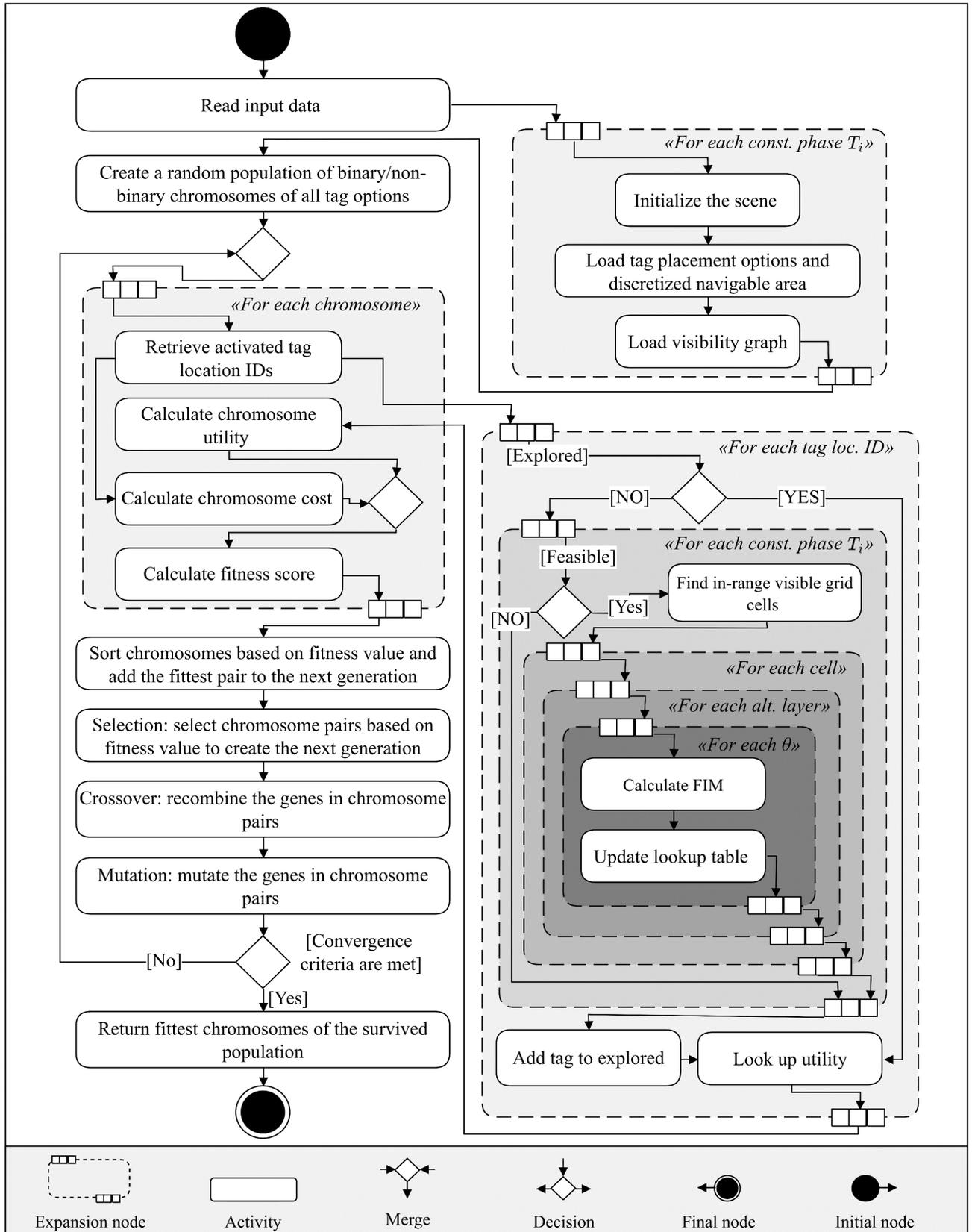

**Figure 7 -** UML activity diagram of GA Engine.

## VALIDATION

The proposed method was implemented in Python 3.8 with process-based parallelism. A sample model of a $64$-$m^2$ unit in its early construction stages was chosen as an exploratory case. As depicted in Figure 8, the project has three phases with identical ROIs. Moreover, a project of size $470$-$m^2$ (Figure 10) was incorporated to further investigate the proposed method's capability in handling large projects. The experiments were carried out in a BIM-enabled, photo-realistic simulation environment, enabling safe and efficient tests supported by absolute ground truth data. The simulation tool was developed on top of *Parrot-Sphinx* ("Parrot-Sphinx 1.2.1" n.d.) and *Gazebo,* the open-source 3D robotics simulator in the Robotic Operating System (ROS) (Quigley et al. 2009). It was shown that the simulation experiments accurately mimic real-world experiments conducted in controlled laboratory environments (Kayhani et al. 2022). The platform used in the experiments was *Parrot Bebop2*, a compact, inexpensive UAV. This section addresses the following questions: (1) How effectively does the proposed method maximize localizability and enhance robustness? (2) How do different metric functions perform? (3) How well can the proposed cost function avoid unnecessary changes in the 4D tag placement of a multi-phase project? (4) What are some suggested values for the main parameters in *PGA-TaPP*? (5) How well does the method scale up to large projects?

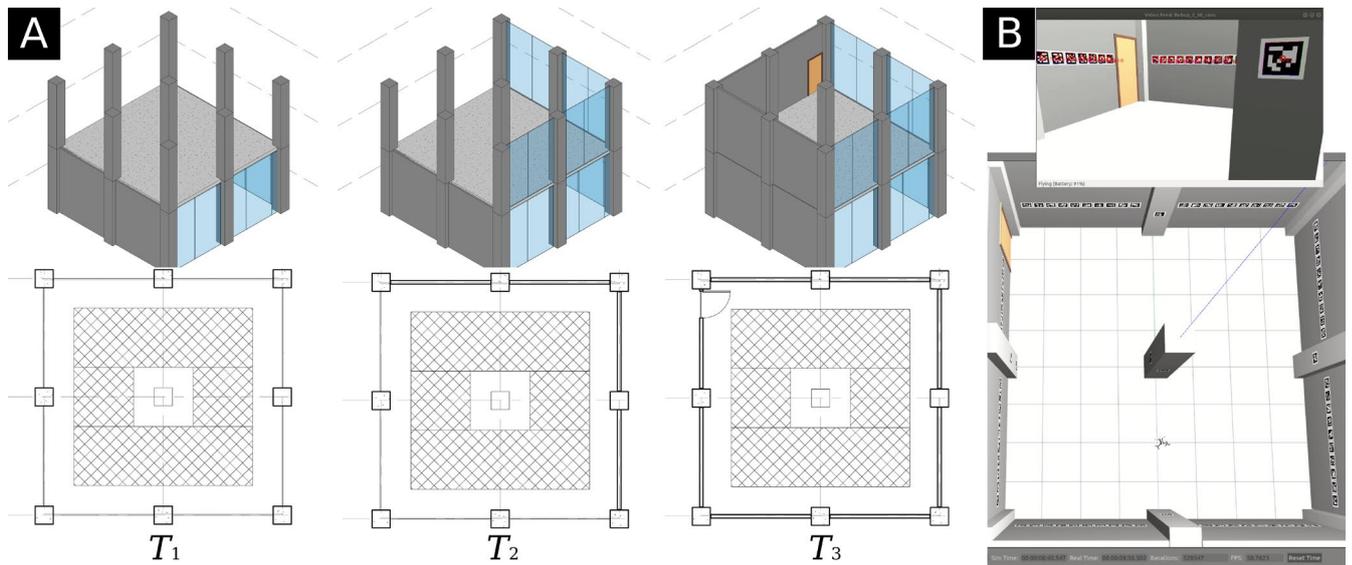

**Figure 8 –** (A) Sample project with three phases (top); the plan view with ROIs hashed and (bottom) 3D models; (B) The UAV in the simulation environment (bottom) and its camera image stream with detected tags (top).

*Experiment Setup*

**Localizability Experiments**

The root-mean-square error (RMSE) in 3D position estimates in different scenarios was compared to investigate the effectiveness of the proposed utility and candidate metric functions in maximizing localizability. The optimal configuration was found in each experiment based on a certain maximum number of tags and different metric functions, including FIM- and non-FIM-based. Based on previous observations (Kayhani et al. 2020), larger tags in the image often result in more accurate localization. To investigate this hypothesis, the area of the detectable tags in the image was incorporated as an alternative to the FIM to compute utility. A semi-random algorithm (denoted as *random*) was implemented as a baseline for comparison. The *random* algorithm generates 100 feasible tag configurations and returns the one with the highest score. Finally, the configuration in which all tag placement options are occupied determines the RMSE's lower bound for tag-based localization. Due to the limited camera FOV and line-of-sight, the measurements' quality and quantity depend on the vehicle's pose. Therefore, the RMSE may differ depending on the UAV's trajectory. Accordingly, we limited our investigations to three planar trajectories that are common maneuvers in automated indoor construction data collection: (1) *Crab Walk* (CWK); (2) *Look Straight Ahead* (LSA); and (3) *Spinning* (SPN) (Figure 9).

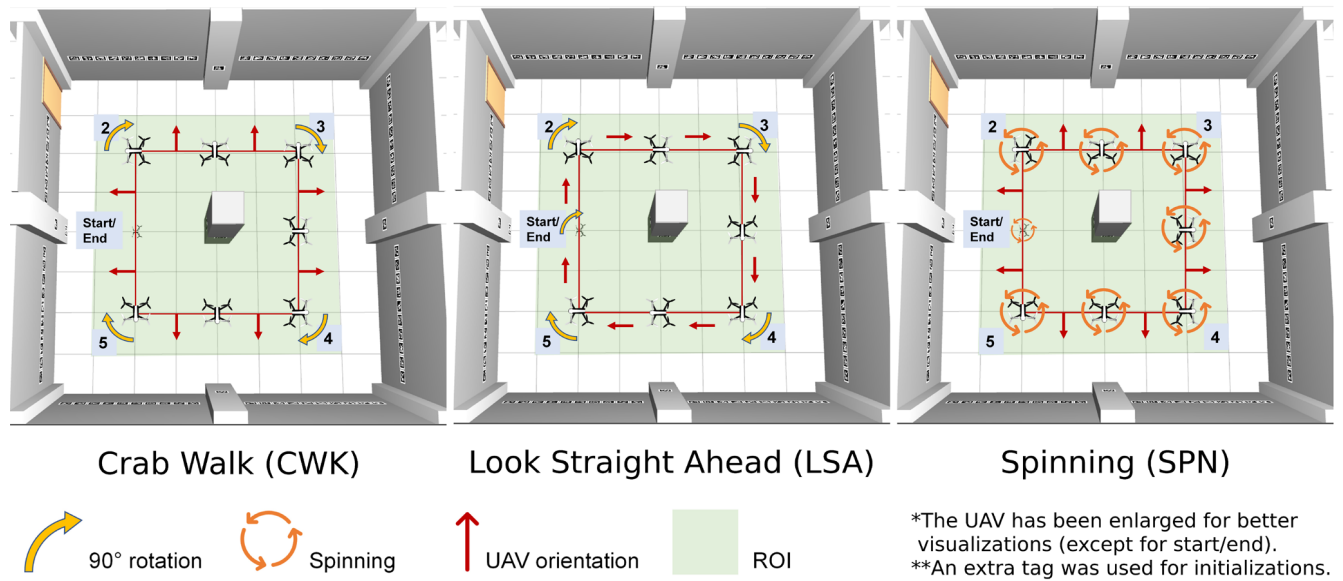

Crab Walk (CWK)  Look Straight Ahead (LSA)  Spinning (SPN)

90° rotation  Spinning  UAV orientation  ROI

*The UAV has been enlarged for better visualizations (except for start/end).
**An extra tag was used for initializations.

**Figure 9** – The threefold trajectories used in the localizability experiments.

The localizability experiments consider a scenario with a single phase, flight altitude, tag installation height, and tag size. In these experiments, the project's 3D BIM was imported into the simulation environment, where tag models were automatically generated and populated based on the 85 tag placement options identified in *Modeling* *and Problem Formulation.* The UAV autonomously followed the paths in Figure 9, where the odometry and the camera feed were recorded for further analyses. The 3D position of the UAV was estimated using our tag-based localization, given the tag configuration obtained in each scenario. These tests concentrate on phase $T_3$ in Figure 8, where the flight altitude and tag installation height were set to 1.5 $m$ and only tags of size 23 $cm \times 23$ $cm$ were assumed to be available.

**Cost Function Experiments**

An optimal 4D tag placement plan uses more practical tag sizes while providing maximum localizability with a minimum number of tags and tag configuration adjustments between phases. The installation costs are regularized by penalizing the deployment of less desirable tag sizes and changes in tag configuration between the project phases. The cost function experiments aim to study the effectiveness of our approach in reducing installation costs. To this end, the same multi-phase project shown in Figure 8 was used. Flight altitudes included 1.5 $m$ and 2 $m$, the tag installation heights were 1 $m$ and 1.5 $m$, and tags with side lengths of 23 $cm$, 16.5 $cm$, and 12.5 $cm$ were assumed to be available. In these experiments, localizability utilities and the number of placements and removals were compared with and without the cost function incorporation. Unlike the localizability experiments, the maximum number of tags was 32. However, with the cost function incorporated, the optimal number of tags in each phase was found also based on the minimum tag score contribution, specified by $w_{plc}$.

**Table 2 -** *PGA-TaPP* Parameters.

| Parameter | Value |
|---|---|
| ROI discretization resolution [m] | 0.5 |
| Rotation discretization step ($\Delta\theta$) [deg] | 20 |
| Tag placement discretization resolution [m] | 0.3 |
| UAV's camera depth of view (DOV) [m] | 8.0 |
| ROIs' importance factor ($I_{roi}$) | 1.0 |
| Population size | 50 |
| Maximum number of iterations | 5000 |
| Mutation function | Flip |
| Crossover function | Single-point |

**Large Project Experiment**

To evaluate the proposed method's performance in handling large projects, a $470\text{-}m^2$ project with five phases was chosen (Figure 10). The installation height and flight altitude were limited to 1.5 $m$. The maximum number of tags was 80, and tag sizes were limited to 16.5 $cm$, printable on letter size sheets. In this scenario, ROIs and tag placement options change with the construction progress. New locations may need to be visited by the UAV, and placement options might be created/removed for tag installation. Finally, the planning and GA parameters listed in Table 1 were kept unchanged across all experiments to make the results comparable.

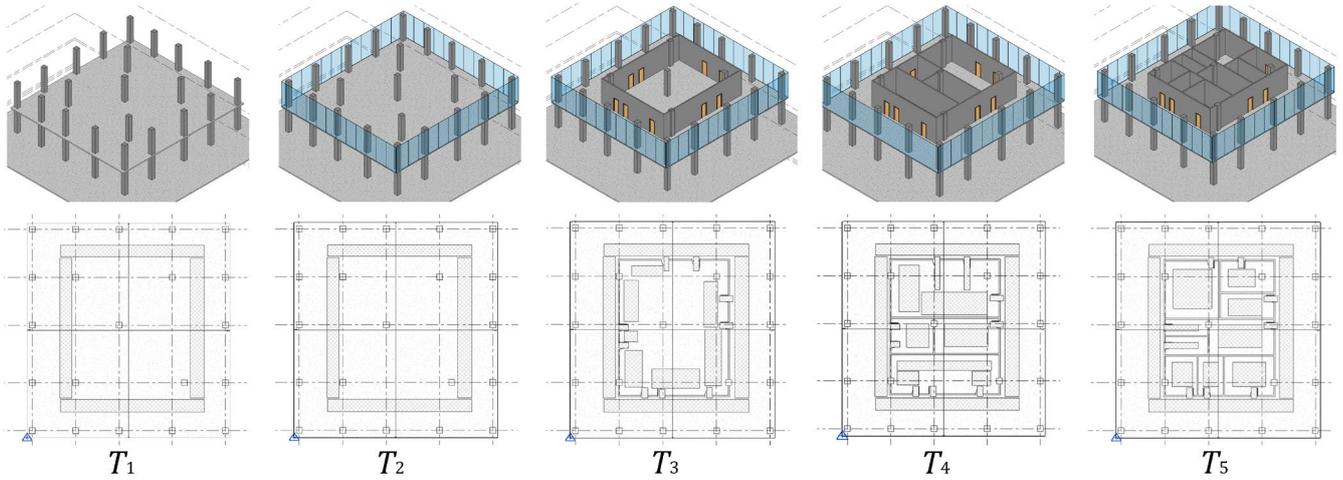

$T_1$          $T_2$          $T_3$          $T_4$          $T_5$

**Figure 10-** A **470-$m^2$** project with five phases. ROIs (hashed polygons) change with construction progress.

***Results***

**Localizability Experiments**

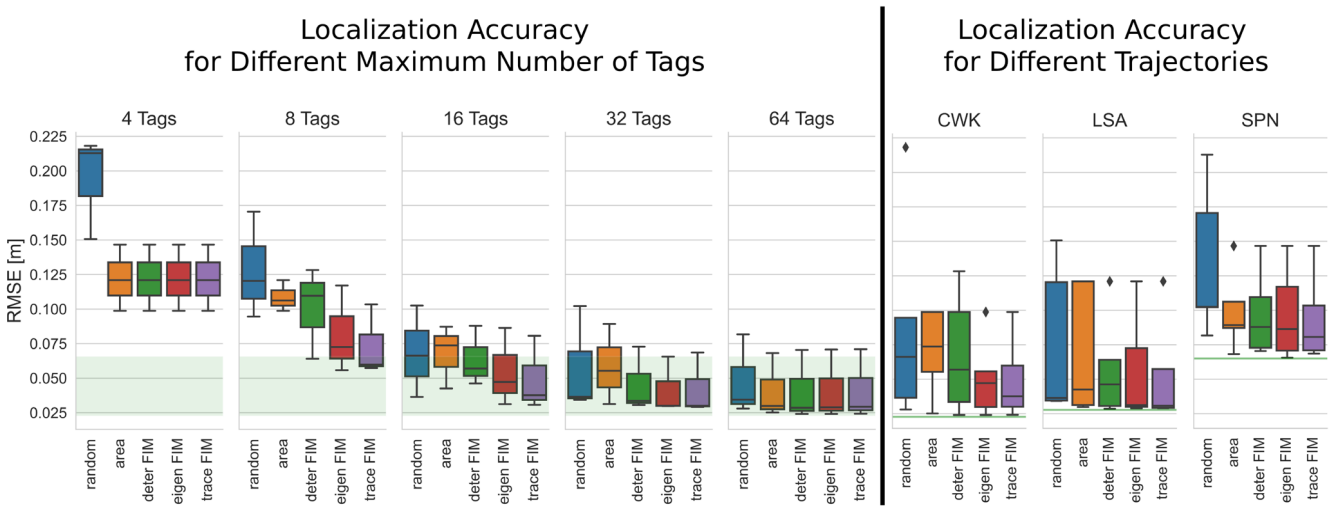

**Figure 11-** The RMSE in 3D position estimates for different metric functions in Localizability Experiments. The green range (left) and line (right) correspond to the RMSE's lower bound, where all tag placement options were occupied.

Figure 11 compares the localization accuracy for the optimal tag configurations in the threefold trajectories, grouped by metric functions and the maximum number of tags. Overall, by adding more tags, RMSE decreases as more measurements are received. Among the trajectories, *SPN* seems to be more challenging for our localization method, as it includes many rotational motions. Almost in all scenarios, the FIM-based utility outperforms the alternative metrics in terms of localization accuracy, among which trace had the lowest error in virtually all scenarios. Therefore, we limit our discussions to the FIM trace as the metric function for the rest of this paper. The optimal tag configurations found using the FIM trace are shown in Figure 12, where the modified ROIs (compare with Figure 9) are illustrated as green boxes, and occupied tag placement options are shown as orange segment lines with an arrow symbol indicating their direction. The heatmap in Figure 12 suggests the smooth distribution of the normalized grid utilities, where green cells are closer to their maximum attainable utility than yellow and red ones.

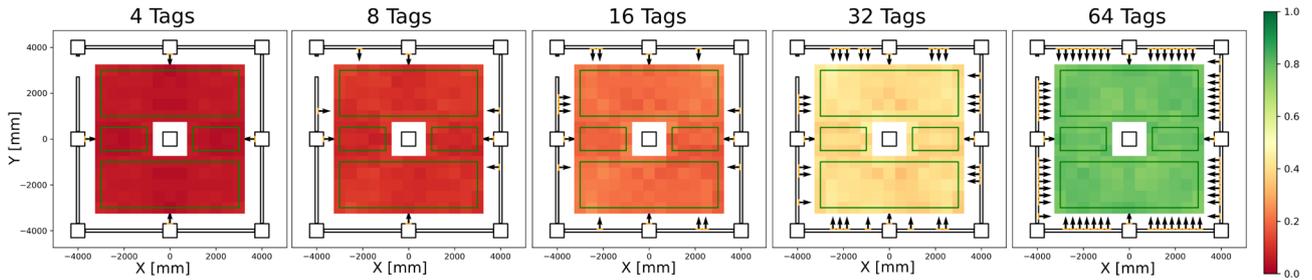

**Figure 12-** The suggested tag configurations in Localizability Experiments using the FIM trace. The modified ROIs are displayed as green boxes, and the normalized cell utilities are visualized as a heatmap.

## Cost Function Experiments

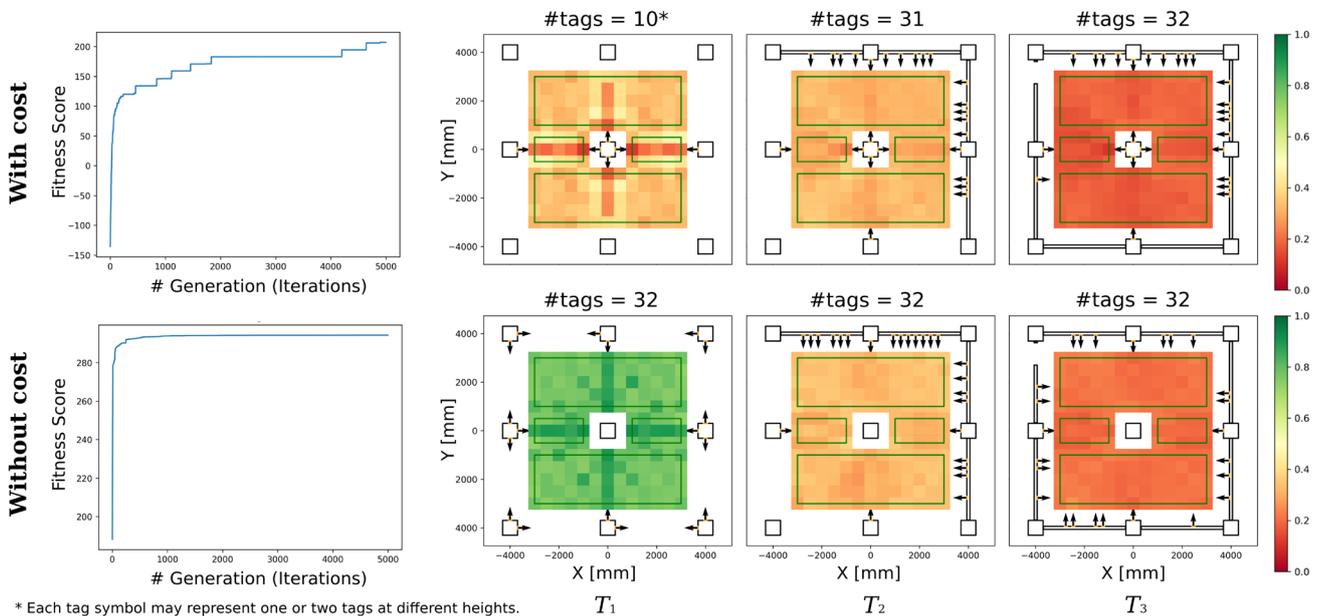

* Each tag symbol may represent one or two tags at different heights.

**Figure 13-** Impact of cost function on penalizing tag configuration changes throughout the project. Left: the fittest solution score in each iteration. Right: the optimal tag configurations in each phase.

As shown in Figure 13, the GA increases the fitness score with more iterations and successfully converges to a solution. Without the cost function, the algorithm finds the best tag locations for each phase separately. However, there is no mechanism except for the maximum number of tags to limit placing more tags per phase. The algorithm can choose between tag sizes based only on the utility. More importantly, as there is no notion of time and the relationship between the phases in the utility, the algorithm repeatedly changes the location and the size of the tags between phases. Without cost, 70 tags were used throughout the project. With cost function, on the other hand, *PGA-TaPP* optimizes the tag configuration by using 33 tags while mainly using the more practical reference tag (16.5 $cm$) and keeping the changes in the network at a minimum. The results and the input parameters are summarized in Table 3.

**Table 3 –** Input parameters and the number of placement and removals in Cost Function Experiments.

| | | Cost Function Experiments | | Large Project Experiment |
|---|---|---|---|---|
| | | With cost | Without cost | |
| Input Parameters | $w_{plc}$ | $S_{min}$= 6%, $P_c$= 2% | 0 | $S_{min}$= 1%, $P_c$= 0.5% |
| | Tag sizes [cm] | [12.0, 16.5, 23.0] | [12.0, 16.5, 23.0] | [16.5] |
| | $\alpha_i$ | [0.5, 1.0, 0.5] | N/A | [1.0] |
| | $\lambda_{rmv}, \lambda_{rpl}$ | 0.1, 0.0 | N/A | 1.0, 0.0 |
| Outputs | $n_{plc}$ (total) | **[2, 26, 5] (33)** | [21, 16, 33] (70) | [142] |
| | $n_{rmv}$ | **1** | 28 | 70 |

**Large Project Experiment**

The previous experiments qualitatively and quantitatively showed that *PGA-TaPP* could successfully maximize localizability via the utility function while minimizing the installation costs through the cost function. Although the limited tag placement options differed between the phases, the ROIs were assumed to be unchanged in the previous experiments. This experiment investigated the method's scalability by studying a more sophisticated scenario with almost 1000 tag placement options across phases where ROIs update as the project progresses.

As seen in Figure 14, the larger search space makes the convergence slower. However, given the input parameters in Table 3, *PGA-TaPP* provides a 4D plan with a minimal number of tag adjustments throughout the project. As shown in Figure 14 and Figure 15, the number of tags in $T_1$ and $T_2$ were 14 and 12, respectively, and less than 80. This is to limit the number of removals in the later phases. However, from $T_3$ on the maximum number of tags were deployed. Figure 15 illustrates the number of tags previously or newly added to the network in each phase.

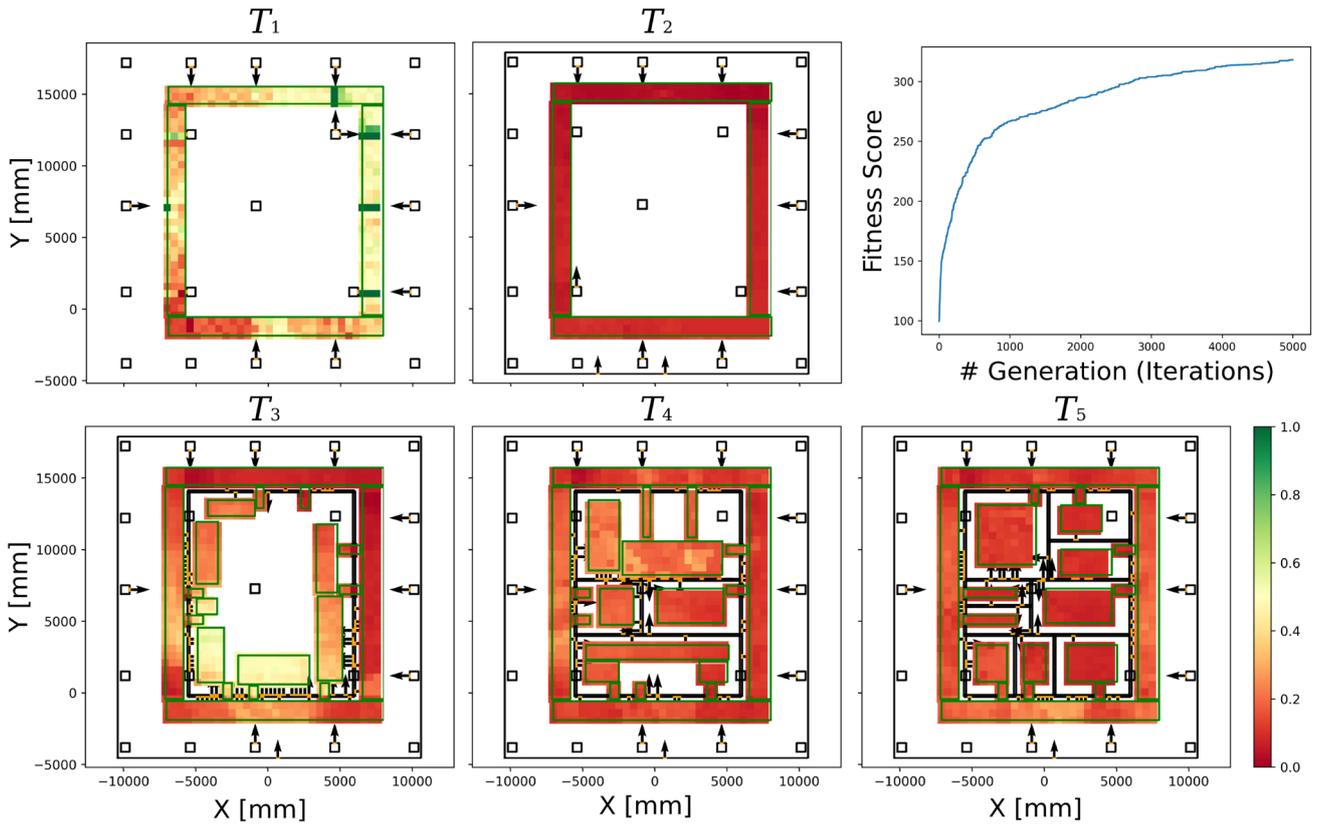

**Figure 14 –** The 4D tag placement plan for the large project where the ROIs and layout change in different phases.

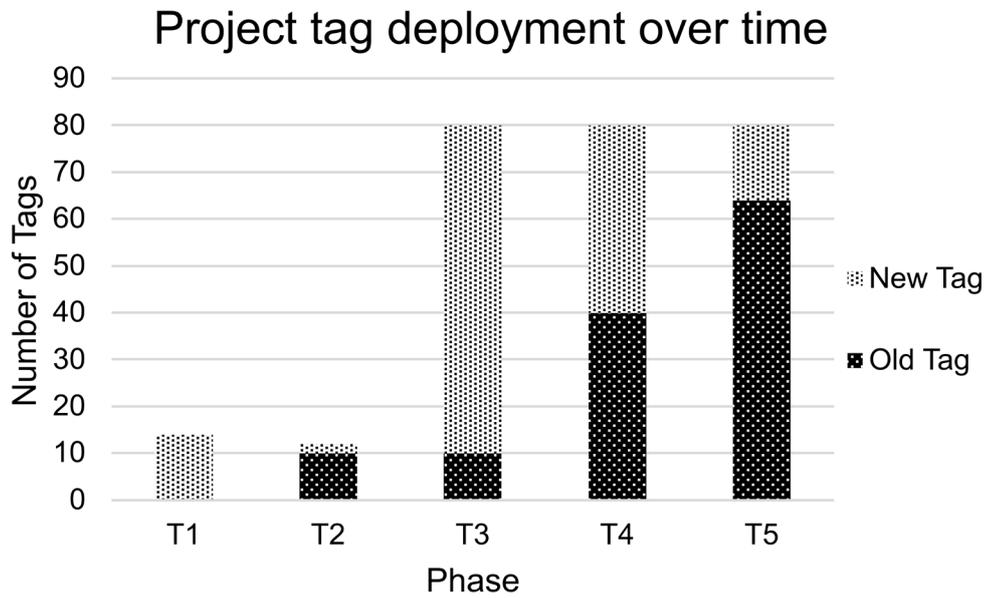

**Figure 15 –** Project tag deployment over time for Large Project Experiment.

*Discussion*

**Localizability with a minimum number of tags**

To reduce the manual tag placement, the planner needs to maximize localizability with a minimum number of tags. The results in Figure 11 indicate that the FIM and the candidate metric functions effectively capture the notion of tag-based localizability, unlike area and the *random* algorithm. Above all, the trace FIM achieved the RMSE's lower bound, i.e., all tag placement options were occupied, using only 40% of the available options. This demonstrates that the proposed utility function effectively maximizes localizability and enhances robustness by smoothly distributing the utility across the grids. Additionally, the trace was shown to perform better than the other more computationally expensive FIM-based metric functions, which can also improve planning efficiency.

**Construction progress, project schedule, and tag size**

The indoor layout changes with construction progress, requiring the planner to take the dynamic nature of construction into account and avoid any unnecessary tag network adjustments as the project progresses. To this end, the proposed method automatically extracts the indoor 2D layout at given times from the as-designed or updated 4D-BIM. Without a 4D-BIM, this information should be extracted manually. The results from Cost Function Experiments (Figure 13) suggest that by incorporating the proposed cost function, *PGA-TaPP* considers the entire project to place tags with higher size desirability in locations that would contribute to the localizability in multiple phases. As summarized in Table 3, with cost function incorporation, 33 tags were used for the whole project, suggesting a 57% reduction in the number of tags without cost function incorporation. The results suggest that the cost function successfully minimized the tag network modification and installation expenses. The algorithm tends to keep the tags for multiple phases, which may result in using fewer tags in the earlier phases. Although this behavior can be modified using the input parameters, we observed that the suggested parameters summarized in Table 2 work well for most scenarios.

**Scalability**

A practical 4D tag placement planner must remain relevant in large-scale construction projects. The results in Figure 14 qualitatively demonstrate the efficacy of the proposed method in handling large projects. The modification of the tag network was kept to a minimum, while the localizability was maximized, despite the change in ROIs and layout between phases. The heatmap visualization is based on the capacity utilization of each cell. In this case, the number of tag placement options increases with construction progress, while the maximum number of tags available is assumed to be 80. Therefore, the cell capacity increases more drastically than the utility gained, resulting in red cells. As shown in Figure 15, *PGA-TaPP* plans the tag placement so that the tags from previous phases can be used in the next ones while maximizing the localizability with a limited number of tags. The total number of tags used was 14.8% of the total number of the options, which proves that the proposed method can effectively optimize the size, location, and number of tags to reduce the installation efforts and costs.

## CONCLUSION

The tag-based visual-inertial method addresses some of the most important technical and practical localization challenges in indoor construction environments. These challenges include frequent layout changes, feature scarcity, and perceptual aliasing. However, tag placement and maintenance could be tedious if not properly planned. This study presented *PGA-TaPP,* a perception-aware genetic algorithm-based tag placement planner, to automatically identify the optimal tag configuration (i.e., size, location, number) and support tag-based visual-inertial indoor localization. This method maximizes the localizability while minimizing the tag installation costs by incorporating the project schedule and safety requirements. It considers multiple project phases, tag placement heights, flight altitudes, and metric functions. Localizability is quantified using the Fisher information matrix (FIM), and the installation cost is kept at a minimum by penalizing tag network modifications (e.g., adding extra tags). The effectiveness of our method was quantitively and qualitatively shown via three case studies. It was demonstrated that the proposed tag-based formulations for estimating the observed FIM represent localizability effectively. It was also experimentally demonstrated that the FIM trace could be a better metric function than determinants and the minimum eigenvalue. We showed that the maximum localizability was obtained using FIM-trace deploying only 40% of the available tag placement options in a single construction phase. In a sample multi-phase project, the proposed cost function could reduce the number of deployed tags by 57% throughout the project. Finally, our experiments confirmed that the proposed method was scalable to larger projects, where the optimal number of tags was limited to 14.8% of the total number of options.

This work contains some limitations. First, even though using metric functions can help compare FIMs, it cannot solely capture the spatial distribution of scores within and among ROIs. For example, the optimization algorithm may maximize the utility in scattered cells or isolated areas. Although robust localization is satisfied in these regions, reliable autonomous navigation will not necessarily be guaranteed between these isolated areas. These areas with locally concentrated high scores, denoted as *islands*, must be avoided. Although our experiments showed a relatively smooth distribution of scores, more studies are needed to comprehensively address this issue. As a suggestion, an alternative policy may use a maximum threshold for each metric function to distribute the optimization attention and avoid *islands*. This threshold implies that we are no longer concerned with improving the localization quality once it reaches a certain level. Second, despite the intuitiveness and flexibility of evolutionary algorithms such as GA, they are local search techniques and do not guarantee global optimums. GA-based optimization can be slow in searching large spaces despite our efforts to parallelize processes. Accordingly, experimenting with other optimization techniques is highly recommended. Third, this work was validated in a simulation environment, mimicking controlled laboratory settings, as a proof of concept, while construction sites are dynamic and cluttered. A complementary study is needed to investigate the performance of the proposed method on-site during construction. Finally, other strategies for tag distribution, including surface spraying, using reflective plastic sheets, e.g., vinyl sheets used in street signs, and horizontal tag placement, e.g., on the floor, should be studied. Other suggestions for future work include investigating the impact of on-site dynamic (e.g., workers) and temporary objects (e.g., formworks), dust, and lighting conditions on localizability.

## DATA AVAILABILITY STATEMENT

All data, models, or code that support the findings of this study are available from the corresponding author upon reasonable request.

## ACKNOWLEDGMENTS

The authors appreciate the financial support from the Natural Science and Engineering Research Council (NSERC) grant number RGPIN-2017-06792. The opinions, findings, and conclusions presented in this work are those of the authors and do not necessarily reflect the views of the entity mentioned above.